\documentclass{llncs}
\usepackage{makeidx}  

\usepackage{amssymb}

\usepackage{amsmath}

\usepackage{lipsum}
\usepackage[utf8]{inputenc} 

\usepackage{lineno}

\usepackage{hyperref}

\usepackage{multirow}

\usepackage{xargs}                      
\usepackage[pdftex,dvipsnames]{xcolor}

\usepackage[colorinlistoftodos,prependcaption,textsize=tiny]{todonotes}

\newcommandx{\unsure}[2][1=]{\todo[linecolor=red,backgroundcolor=red!25,bordercolor=red,#1]{#2}}
\newcommandx{\change}[2][1=]{\todo[linecolor=blue,backgroundcolor=blue!25,bordercolor=blue,#1]{#2}}
\newcommandx{\info}[2][1=]{\todo[linecolor=OliveGreen,backgroundcolor=OliveGreen!25,bordercolor=OliveGreen,#1]{#2}}
\newcommandx{\improvement}[2][1=]{\todo[linecolor=Plum,backgroundcolor=Plum!25,bordercolor=Plum,#1]{#2}}
\newcommandx{\thiswillnotshow}[2][1=]{\todo[disable,#1]{#2}}

\usepackage{color}
\usepackage{graphicx}

\usepackage{wrapfig} 

\usepackage[linesnumbered,ruled]{algorithm2e}
\usepackage[noend]{algpseudocode}

\begin{document}
\title{A Hierarchical Two-tier Approach to Hyper-parameter Optimization in Reinforcement Learning}
\titlerunning{A Hierarchical Two-tier Approach to Hyper-parameter Optimization in Reinforcement Learning}  
%
\author{Juan Cruz Barsce\inst{1} \and Jorge A. Palombarini\inst{1, 2, 3} \and Ernesto Martínez\inst{4}}
\authorrunning{Barsce et al.} 
%
%
\institute{Dpto. de Ingeniería en Sistemas de Información,  Facultad Regional Villa María, UTN, Argentina
\and
GISIQ, Facultad Regional Villa María, UTN, Argentina
\and
CIT Villa María - CONICET-UNVM, Argentina
\and
Instituto de Desarrollo y Diseño CONICET-UTN, Argentina,\\
\email{ecmarti@santafe-conicet.gob.ar}
}

\maketitle              

\begin{abstract}
  Optimization of hyper-parameters in reinforcement learning (RL) algorithms is a key task, because they determine how the agent will learn its policy by interacting with its environment, and thus what data is gathered. In this work, an approach that uses Bayesian optimization to perform a two-step optimization is proposed: first, categorical RL structure hyper-parameters are taken as binary variables and optimized with an acquisition function tailored for such variables. Then, at a lower level of abstraction, solution-level hyper-parameters are optimized by resorting to the expected improvement acquisition function, while using the best categorical hyper-parameters found in the optimization at the upper-level of abstraction. This two-tier approach is validated in a simulated control task. Results obtained are promising and open the way for more user-independent applications of reinforcement learning.

\keywords{reinforcement learning, hyper-parameter optimization, Bayesian optimization, Bayesian optimization of combinatorial structures (BOCS)}
\end{abstract}

\section{Introduction}
Generalizing from data in a machine learning algorithm involves a training process, where such algorithm learns the model structure and parameters that best fit the available data. Training, in turn, depends on a prior design process that defines the hyper-parameters that constraint the conditions of data-driven learning. Setting them properly is crucial to the learning process, and can make the difference between mediocre and state-of-art model prediction \cite{hutter_beyond_2015}. In particular, optimizing the hyper-parameters of reinforcement learning algorithms  \cite{sutton_reinforcement_2018} is a hard task, because data is not provided \textit{a priori}, but increasingly generated through interactions with the environment. Hence, hyper-parameters determine which data is generated. In turn, such data determines the parameter values, which also influences the next set of data generated, and so on and so forth.

Hyper-parameters are usually manually optimized, which can be very inefficient \cite{hutter_beyond_2015}, or by methods such as random search \cite{bergstra_random_2012} or Bayesian optimization (BO) \cite{mockus_application_1978} \cite{shahriari_taking_2016}. The latter performs a black-box optimization of a function $f$, resorting both to a prior distribution $f \sim P(y)$ and to the data available points $(X, y)$ in order to compute the mean and variance for unseen inputs, typically predicted by Gaussian process (GP) regression \cite{rasmussen_gaussian_2008}, which is used to maximize an \textit{acquisition function} that is cheap to optimize globally. The most common acquisition function is \textit{expected improvement}, where the next point is decided by considering the probability of the next maximum, pondered by the predicted variance. The issue with random search is that the method uses very limited information about previous queries of $f$. On the other hand, while Bayesian optimization uses information regarding past queries, it also has two limitations similarly to random search: 1) it involves no assumption about the influence of hyper-parameters on the information content, and 2) it is no efficient to optimize categorical hyper-parameters such as the RL algorithm selected. In this work, an algorithm is proposed that, by assuming a hierarchical relationship between RL hyper-parameters, optimizes such structural hyper-parameters first, and then uses traditional Bayesian optimization to tune the real-valued hyper-parameters of the learning algorithm. The proposed algorithm is validated against random search and Bayesian optimization in the classical Cart-pole environment.

\section{Reinforcement learning}

Reinforcement learning \cite{sutton_reinforcement_2018} is a sub-area of machine learning involving an autonomous agent that must control an external environment while learning a control policy that maximize the received reward from such environment. Formally, it can be stated as a Markov Decision Process, $(S, A, R(.), P(.), \gamma)$, where $S$ is a set of environmental states, $A$ is a set of actions available to the agent, $R(s)$ is an external function that assigns the agent a reward to state transition caused by the agent action taken at any state $s$, $P(s' \mid s,a)$ is a function that determines the probability that the agent transitions from a state $s$ to a state $s'$ when the action $a$ is taken, and finally, $\gamma \in [0,1)$ is a real number that discounts the values of future rewards.

The control policy is defined as a function $\pi(a \mid s)$, and represents the probability of taking the action $a$ when the environment is in state $s$. With $\pi(.)$, the agent aims to maximize the \textit{value function} for every state, defined as the expected reward starting from a given state at time-step $t$ and following a given policy $\pi$ thereafter. Formally such function must satisfy the Bellman equation \cite{sutton_reinforcement_2018}. A crucial aspect in RL is the trade-off between exploration and exploitation, in which the agent has to choose between taking actions that are considered to be the best according to the current estimation of the optimal policy learned, or taking actions that are deemed as sub-optimal but makes room for the agent to discover better actions to exploit in the future.

Among basic RL algorithms the most commonly used are \textit{Q}-Learning \cite{watkins_q-learning_1992} and \textit{SARSA} \cite{rummery_-line_1994}. Both algorithms compute the action-value function $Q(s,a)$ according to a temporal difference between the discounted value of $Q(s',a')$ of the next state and action, and the $Q$-value for the current state and chosen action. The difference between \textit{Q}-Learning and \textit{SARSA} is how they choose the next action $a'$, where the latter selects the action based on the policy $\pi$, and thus it is an \textit{on-policy} algorithm, whereas the former selects the best estimated action $a'$ for the resulting state $s'$, therefore it is considered as \textit{off-policy}. Algorithms may also update the $Q$ values of past states and actions that were responsible for reaching the current state, using a mechanism known as \textit{eligibility traces} \cite{sutton_reinforcement_2018}.

To balance exploitation and exploration in this work, the $\epsilon$-greedy policy is used, where the best action $a'$ is chosen with an $1-\epsilon$ probability, and the other alternative actions are chosen at random with a low probability $\epsilon$. Alternatively, the \textit{Softmax} policy is used, where each action is selected based on the equation $\pi(a \mid s) = e^{Q(s,a)/\tau}/\sum_{a'}e^{Q(s,a')/\tau}$, where $\tau$ is an hyper-parameter defines the influence of the $Q$ values in defining the action selection probabilities.

Each of the RL algorithms and policies have their own set of hyper-parameters that must be defined before the agent learning curve begins. Common hyper-parameters includes a learning rate that determines the speed of the convergence of the agent $\alpha \in (0,1)$, an exploration rate $\epsilon \in (0,1)$ if the policy is $\epsilon\textit{-greedy}$, and a discount factor $\gamma \in (0,1)$ for future rewards (see \cite{sutton_reinforcement_2018} for a more detailed description). If the policy used is $\epsilon$-greedy, an additional hyper-parameter known $\epsilon$-decay rate can be used, that reduces the value of $\epsilon$ after an episode, in order to lower the exploration rate of the agent after a given number of episodes has been experienced.

\section{Two-tier hierarchical Bayesian optimization of RL hyper-parameters}

In this work, a method that employs Bayesian optimization to perform a two-tier optimization of both structural and solution-level hyper-parameters of an RL agent is proposed. The objective function proposed in this work, $f : \Theta_a \cup \Theta_s \to \mathbb{R}$ maps a set $\Theta$ of both structural and solution level hyper-parameters of the algorithm to a real number that measures the overall performance of the learning agent, assuming an episodic task. A novel aspect of this approach is that it combines Bayesian optimization for both categorical and real-valued hyper-parameters. For the categorical hyper-parameters, Bayesian optimization of discrete structures (BOCS) \cite{baptista_bayesian_2018} was used, where the categorical hyper-parameters are taken as binary variables and the maximum of the acquisition function is found through simulated annealing, instead of relying on Gaussian process regression \cite{baptista_bayesian_2018}. On the other hand, the RLOpt \cite{barsce_towards_2018} approach  was used for the real-valued hyper-parameters. As the distribution of the $f$ function is unknown, the prior assumption is that it follows a multivariate Normal distribution with a mean vector $\mu_0$ and a co-variance function $\Sigma$. In order to calculate the value of $f(\Theta)$ for $\Theta = \Theta_s \cup \Theta_a$, an RL agent is instantiated in a certain environment with hyper-parameters $\Theta$, and it is set to run for a certain number of episodes in order to learn a policy to behave in such a way to maximize its received reward. Whenever the agent is assigned a new $\Theta$ vector, it resets all its prior knowledge about the policy in order to make a fresh start, unbiased by the prior hyper-parameter settings. The instance where an RL agent runs a certain number of episodes under the same hyper-parameter setting is called a \textit{meta-episode}.

This method involves the assumption that RL parameters are related in a two-level hierarchy that takes into account their levels of abstraction. In such relationship, algorithm hyper-parameters such as the exploration policy (e.g. Softmax or $\epsilon$-greedy \cite{sutton_reinforcement_2018}) are in a higher level of abstraction than the solution-level hyper-parameters (e.g. the temperature $\tau$ or the exploration rate $\epsilon$), by the fact that the former establishes the possible values for the latter. Following such an assumption, in the proposed method the structural algorithm hyper-parameters are optimized first, while using a set of prior algorithm hyper-parameters, and storing the pairs $\Theta, f(\Theta)$ of point and its corresponding $f$ output in the initial set $D_1$. Once a certain number of meta-episodes are elapsed, the best structural hyper-parameters are kept frozen and the optimization of the hyper-parameters dependent on such algorithms is started, storing its results in the set $D_2$. The method for such optimization is stated in Algorithm \ref{alg:BO_rl}.

\begin{algorithm}
\caption{Bayesian optimization applied to both the structural and algorithm hyper-parameters}
\label{alg:BO_rl}
\small
	\SetKwInOut{Input}{Input}
	\SetKwInOut{Output}{Output}

  \Input{set of prior algorithm hyper-parameters $\Theta_p$}

  \For{$\text{structural evaluation} = 1$ to N}{
  	Obtain $\Theta_n$ that optimizes $\alpha_{BOCS} (.) \to \mathbb{R}$ over the structural hyper-parameters, with prior algorithm hyper-parameters preset as $\Theta_p$\\

  	Query the objective function $f$ at the point $\Theta_n = \Theta_s \cup \Theta_p$ \\

  	$D_1 \gets D_1 \cup \{\Theta_n, f(\Theta_n)\}$\\

  	Update the model
  }

  Initialize $D_2 \gets \{\Theta^+, f(\Theta^+)\}$ with the best $\Theta^+ = \Theta_s^+ \cup \Theta_p$ point and corresponding maximum of $f$ found \\

  \For{$\text{hyper-parameters evaluation} = 1$ to M}{
    Obtain $\Theta_m = \Theta_s^{+} \cup \Theta_a$ that optimizes $\alpha_{EI}(.) \to \mathbb{R}$, with structural hyper-parameters preset as $\Theta_s^+$ \\

    Query the objective function $f$ at the point $\Theta_m = \Theta_s^{+} \cup \Theta_a$ \\

    Add the result $f(\Theta)$ to $D_2$\\

    Update the statistical model (e.g. Gaussian process)
  }

  \Return{$(\arg\max_\Theta f, \max f)$}

\end{algorithm}

\section{Computational experiments}


The proposed approach is validated in a discretized version of the classic Cart-pole control environment, which consists of an environment with a cart that moves either left or right, and it is holding a pole that can swing in both directions. The objective for the cart is to keep the pole balanced (i.e. by not letting it in a position where it will fall to the floor), while maintaining itself within certain limits. Each episode is terminated whether the pole position is above or below 12 degrees from the vertical position, when the cart moves beyond a distance of 2.4 units from the center, or when 200 time-steps have elapsed. A reward of +1 is given after every time-step when the pole is still maintaned upright, and a reward of -200 is assigned to the agent whenever the pole has fallen. The implementation used for the environment was the OpenAI Gym implementation \cite{brockman_openai_2016}.

The proposed Algorithm \ref{alg:BO_rl} is compared against two of the most common methods for hyper-parameter tuning: random search and Bayesian optimization. To optimize RL hyper-parameters with the latter, the RLOpt framework \cite{barsce_towards_2018} is used. A total number of 30 meta-episodes were used for the three approaches, where the average reward was used to compute $f$ on each meta-episode. In the proposed algorithm, 10 meta-episodes were used to optimize the discrete hyper-parameters and 20 meta-episodes were used to optimize the real-valued hyper-parameters once the structural hyper-parameters were fixed. The structural hyper-parameters optimized were $\textit{algorithm} \in \{Q\textit{-learning}, \textit{SARSA}\}$, $\textit{eligibility-traces} \in \{true, false\}$, $\textit{policy} \in \{\epsilon\textit{-greedy}, \textit{Softmax}\}$ and $\epsilon\textit{-decay} \in \{\textit{true}, \textit{false}\}$ (it only applies when $\epsilon$-greedy policy is selected). On the other hand, the algorithm hyper-parameters optimized were $\alpha \in (0,1)$, $\epsilon \in (0,1)$, $\gamma \in (0,1)$, the number of bins that divides the cart position and speed, $\textit{n-bins} \in (5, 20)$, and the number of bins used to discretize the pole angle position and its speed, $\textit{n-bins-angle} \in (5, 20)$.

\begin{figure}[!htb]
  \minipage{0.59\textwidth}
    \includegraphics[width=\linewidth]{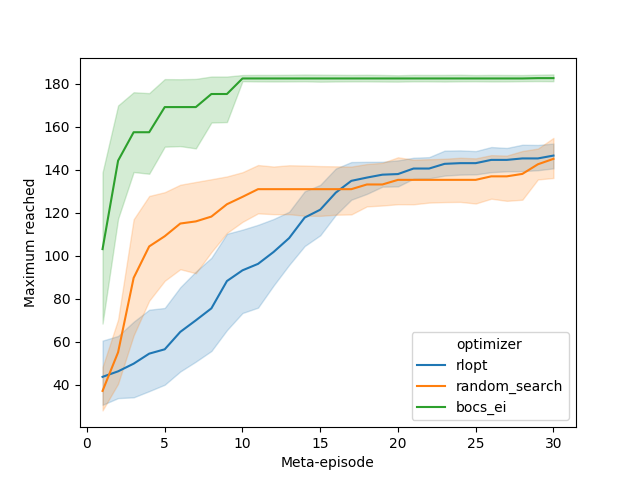}
    \caption{Average maximum reached by each optimizer, per each meta-episode}
    \label{fig:maximum_reached}
  \endminipage\hfill
  \minipage{0.59\textwidth}
    \includegraphics[width=\linewidth]{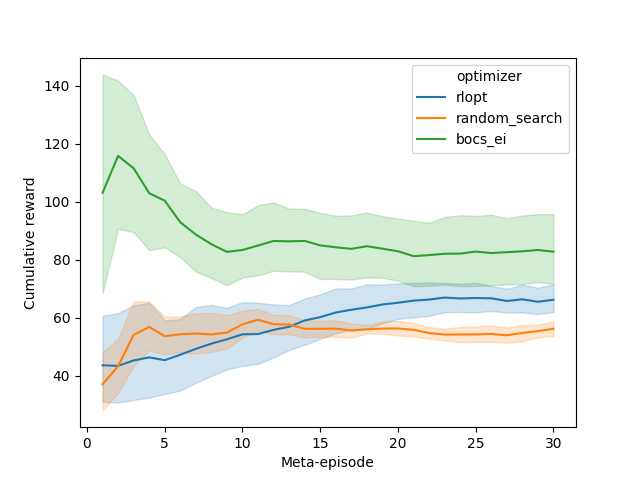}
    \caption{Cumulative reward per meta-episode reached by each optimizer}
    \label{fig:cumulative_reward}
  \endminipage\hfill
\end{figure}

Results obtained are shown in Fig. \ref{fig:maximum_reached} and Fig. \ref{fig:cumulative_reward}, where the thick lines and their nearby curves correspond to the average and the 95\% confidence interval for ten simulations with different random seeds. As can be seen, the proposed method is consistently better at finding the average set of hyper-parameters that reach the maximum than the other two methods that does not optimize the structural hyper-parameters. In Fig. \ref{fig:cumulative_reward}, it can be appreciated that the proposed method starts reducing its average cumulative reward after having the very first initial convergence where the maximum was found for the ten executions. The average execution time was 7, 8 and 12 minutes for the random search, RLOpt, and for the proposed optimizer, respectively.

\section{Concluding remarks}

In this work, a novel approach that involved the optimization of both categorical and real-valued RL hyper-parameters, assuming a hierarchical relationship between them was presented. The validation in the Cart-pole environment highlights that the proposed approach performs consistently better than the monolithic optimization of the real-valued hyper-parameters alone. Our current research efforts are focused on including the extension of the concept of a hierarchical relationship among many hyper-parameters, the optimization of complex computational structures such as deep neural networks, and the use of methods such as power analysis in order to determine whether the sample size of meta-episodes must be increased, among others.

\bibliographystyle{amsplain}
\bibliography{main.bib}

\end{document}